%% file: 0-title.tex
\pdfoutput=1

\documentclass[11pt]{article}

\usepackage{ACL2023}

\usepackage{times}
\usepackage{latexsym}
\usepackage{paralist}

\usepackage[T1]{fontenc}

\usepackage[utf8]{inputenc}

\usepackage{microtype}

\usepackage{inconsolata}

\usepackage{graphicx}
\usepackage[labelfont=bf]{caption}
\usepackage{adjustbox}
\title{VAKTA-SETU: A Speech-to-Speech Machine Translation Service in Select Indic Languages}

\author{{Shivam Mhaskar, Vineet Bhat, Akshay Batheja,}\\{\bf Sourabh Deoghare, Paramveer Choudhary, Pushpak Bhattacharyya} \\[1ex]
CFILT, Indian Institute of Technology Bombay \\
\texttt{\{shivammhaskar, akshaybatheja, sourabhdeoghare, paramc, pb\}@cse.iitb.ac.in}\\
\texttt{\{vineetbhat2104\}@gmail.com}
}

\begin{document}
\maketitle

\input{1-abstract.tex}
\input{2-intro.tex}
\input{3-related-work.tex}
\input{4-method.tex}

\input{5-expt-setup.tex}

\input{6-results.tex}

\input{7-feedback.tex}

\input{8-deploy.tex}
\input{9-conclusion.tex}

\section*{Limitations}
\input{10-limitations.tex}

\section*{Ethics Statement}
Our work aims to develop and deploy a scalable speech-to-speech machine translation system. For training all the ASR, DC, MT, and TTS models, we have used publicly available datasets, and we have cited their sources as well. The training data for NMT, which also includes the data generated by us earlier as a part of another project has already been submitted to concerned agencies which have put the data in the public domain. No user information was present in the datasets protecting users' privacy and identity. Publicly available datasets can sometimes contain biased data. We understand that every dataset is subject to intrinsic bias and that computational models will inevitably learn biased information from any dataset.
 
\bibliography{anthology,custom}
\bibliographystyle{acl_natbib}

\appendix

\section{Appendix}
\label{sec:appendix}

\subsection{Salient Linguistic Properties (Referred from Section \ref{sec:intro})}
\label{app:linguistic}
\subsubsection{English}
English belongs to the Indo-European language family and shows major influences from French and Latin. English has around 600 million native speakers and around 2 billion total speakers. English follows the subject-verb-object (SVO) word order. English has largely abandoned the inflectional case system in favor of analytic constructions. English distinguishes at least seven major word classes: verbs, nouns, adjectives, adverbs, determiners (including articles), prepositions, and conjunctions. English nouns are only inflected for number and possession. English pronouns conserve many traits of the case and gender inflection.
\subsubsection{Hindi}
Hindi belongs to the Indo-Aryan language family and has around 300 million native speakers. Hindi is written in the Devanagari script. Most of the modern Hindi vocabulary is borrowed from Sanskrit. Hindi also has influence from Persian. Hindi follows the subject-object-verb word order. In Hindi, nouns are inflected for number, gender, and case. Hindi has two numbers, singular and plural. It has two grammatical genders, masculine and feminine. And it has two cases direct and oblique. The gender of inanimate objects is not predictable from the form or meaning. Pronouns are inflected for numbers and cases. Adjectives are of two types declinable and indeclinable. Verbs are inflected for person, number, gender, tense, mood, and aspect.
\subsubsection{Marathi}
Marathi belongs to the Indo-Aryan language family and has around 83 million native speakers. Marathi is written in the Devanagari script. Marathi employs agglutinative, inflectional, and analytical forms. Marathi has three grammatical genders, masculine, feminine, and neuter. Marathi follows the subject-object-verb word order. Marathi also shares vocabulary and grammar with Dravidian languages. Marathi follows a split-ergative pattern of verb agreement and case marking. An unusual feature of Marathi, as compared to other Indo-European languages, is that it displays inclusive and exclusive we, common to the Dravidian languages.

\subsection{Background of ASR Systems (Referred from Section \ref{sec:rel_work})}
\label{sec:asr-background}

Speech recognition is often the first task in interactive \& intelligent NLP agents and is a crucial module for downstream systems. However, ASR did not receive much attention till the first half of the $20^{th}$ century. After the 1950s, corporations worldwide started investing in recognition technologies, paving the way for high-quality research and production.

There are 7000+ languages worldwide, but more than half of the world's population uses only 23. Thus ASR has the potential to break the linguistic and communication barriers among the world population. Like any AI system, the amount of data available is significant in designing a state-of-the-art system in ASR. There has been a vast amount of research in popular languages like English, Spanish, and German, known as \textit{high resource} due to the large-scale availability of data and open-source research in these languages. However, \textit{low resource} languages such as Hindi, Marathi, and Tamil do not have high-performance systems due to the lack of transcribed data.

We treat the acoustic input signal as O = \{$o_1$  , $o_2$ , $o_3$ , $o_4$ , \dots \} a series of observations and define a sequence of words as the desired output W = \{$w_1$  , $w_2$ , $w_3$ , $w_4$ , \dots \}.

We would like to get those sequence of words W from the language L, which maximizes the following condition given the acoustic input O.

\begin{equation}
\label{eq-1}
\hat{W} = \arg\max_{W \in L} P(W | O )
\end{equation}

We can use Bayes rule to rewrite this as - 

\begin{equation}
\label{eq-2}
\hat{W} = \arg\max_{W \in L} \frac{P(O|W) P(W)}{P(O)} 
\end{equation}

For every possible sequence of words W, the denominator of equation \ref{eq-2} is the same. Since we are dealing with the argmax operator, we can ignore the denominator and write the final expression as - 

\begin{equation}
\label{eq-3}
\hat{W} = \arg\max_{W \in L} P(O|W) P(W)
\end{equation}

Each component in a ASR system plays an important role in calculating the above two probabilities. 

Word Error Rate (WER) is a common metric used to evaluate ASR systems. Derived from the Levenshtein distance, WER calculates ground-truth deviations at the word level instead of the phoneme level. Word error rate can then be computed as: 

\begin{equation}
\label{eq-4}
WER = \frac{S + D + I}{N} = \frac{S + D + I}{S + D + C}
\end{equation}

where S is the number of substitutions,
D is the number of deletions,
I is the number of insertions,
C is the number of correct words,
N is the number of words in the reference

\begin{table*}[h]
\centering
\begin{tabular}{p{2cm}p{8cm}p{3cm}}
\hline
\textbf{Language} & \textbf{Model} & \textbf{Word Error Rate}\\
\hline
\hline
English & wav2vec 2.0 - base \cite{baevski} & 49.20 \\ 
& wav2vec 2.0 - XLSR \cite{ruder-etal-2019-unsupervised} & 31.57 \\ 
& Whisper - small \cite{Radford2022RobustSR} & 28.40 \\
& wav2vec 2.0 - Vakyansh \cite{DBLP:journals/corr/abs-2107-07402} & 32.80 \\
& wav2vec 2.0 - Vakyansh \& Noisy finetuning & 28.20 \\
\hline 
Hindi & wav2vec 2.0 - XLSR \cite{ruder-etal-2019-unsupervised} &  44.08 \\
& Whisper - small \cite{Radford2022RobustSR} & 34.60 \\
& wav2vec 2.0 - Vakyansh \cite{DBLP:journals/corr/abs-2107-07402} &  19.14 \\
& wav2vec 2.0 - Vakyansh \& Noisy finetuning & 16.19 \\

\hline
\end{tabular}
\caption{Results of evaluating ASR baselines on chosen English \& Hindi test sets. Word Error Rate (WER) is calculated as a percentage and follows an inverse relationship with recognition accuracy. Higher WER indicates a worse model and a lower WER indicates a better model.}
\label{app-tab:res-asr}
\end{table*}

\begin{table*}[h]
\centering
\begin{tabular}{p{2cm}p{5cm}p{1.5cm}p{1.5cm}p{1.5cm}}
\hline
\textbf{Language} & \textbf{Model} & \textbf{Precision} & \textbf{Recall} & \textbf{F1 Score} \\
\hline
\hline
English & MuRIL - SWBD & 94.96 & 94.33 & 94.64 \\ 
& MuRIL - SWBD \& LARD & 97.92 & 95.09 & 96.48 \\ 
Hindi & MuRIL - SWBD &  68.24 & 58.46 & 62.97 \\
& MuRIL - SWBD \& Syn Hi &  85.38 & 79.41 & 82.29 \\

\hline
\end{tabular}
\caption{F1 scores of DC models in English \& Hindi. Adding synthetic disfluent sentences from \cite{passali-etal-2021-towards} improved the performance of our baseline MuRIL transformer \cite{khanuja_muril} trained on the Switchboard corpus \cite{Godfrey1992SWITCHBOARDTS}. We get similiar results in Hindi when adding synthetic disfluent sentences from \cite{kundu-etal-2022-zero} improved the performance over a zero shot baseline.}
\label{app-tab:res-dc}
\end{table*}

\begin{table*}[h]
\centering
\begin{tabular}{p{2cm}p{1.5cm}p{1.5cm}p{1.5cm}p{1.5cm}p{1.5cm}p{1.5cm}}
\hline
\textbf{Test Set} & \textbf{En-Mr} & \textbf{Mr-En} & \textbf{En-Hi} & \textbf{Hi-En} & \textbf{Hi-Mr} & \textbf{Mr-Hi} \\
\hline
\hline
FLORES & 14.67 & 21.67 & 33.15 & 27.74 & 7.43 & 14.77 \\
Tico-19 & 17.11 & 29.10 & 37.70 & 33.04 & 8.45 & 15.88 \\
ILCI & 11.42 & 14.08 & 23.24 & 17.91 & 32.36 & 39.91 \\

\hline
\end{tabular}
\caption{BLEU scores of the NMT models for English-Marathi, English-Hindi, and Hindi-Marathi language pairs on different test sets. We used the sacrebleu \cite{post-2018-call} library to compute the BLEU scores. The \textbf{FLORES} test set consists of 1012 parallel sentences across various domains, so it is a multi-domain test set. The \textbf{Tico-19} test set consists of 2100 sentences from the healthcare domain. The \textbf{ILCI} test set consists of 2000 sentences from the tourism and healthcare domain.}
\label{app-tab:res-mt}
\end{table*}

\begin{table*}[h]
\centering
\begin{tabular}{p{2cm}p{3cm}p{1.5cm}p{1.5cm}p{1.5cm}}
\hline
\textbf{Language} & \textbf{Model} & \textbf{AQ} & \textbf{I} & \textbf{MOS} \\
\hline
\hline
Hindi & Tacotron 2 & 4.63 & 4.21 & 4.42 \\ 
& Forward Tacotron & 4.70 & 4.48 & 4.59 \\ 
Marathi & Tacotron 2 &  4.74 & 4.32 & 4.53 \\
& Forward Tacotron &  4.79 & 4.57 & 4.68 \\

\hline
\end{tabular}
\caption{Results of the subjective evaluation for our TTS systems; Audio Quality (AQ) provides information about the speech quality generated and general prosody of the system output; Interpretability (I) asks surveyors to evaluate the audio outputs for clarity of speech and understanding of the semantic content; All scores are reported out of 5.0}
\label{app-tab:res-tts}
\end{table*}

\subsection{Background of DC Systems (Referred from Section \ref{sec:rel_work})}
\label{sec:dc-background}

\begin{figure}[h]
\centering
\includegraphics[width=75mm]{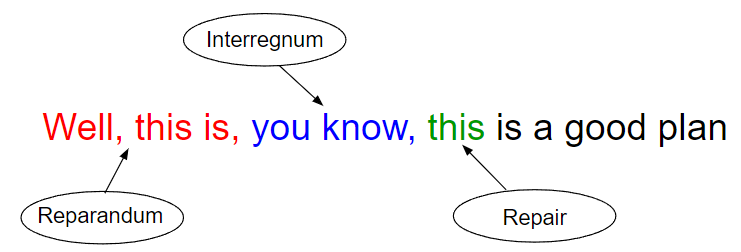}
\caption{Surface structure of disfluencies} \label{fig:dc-example}
\end{figure}

Disfluency correction systems learn the mapping between disfluency structure and types of disfluencies while detecting the presence/absence of disfluent utterances. The structure of any disfluent utterance is composed of three parts - \textit{reparandum}, \textit{interregnum}, and \textit{repair}. The reparandum consists of the words incorrectly uttered by the speaker and will need correction or complete removal. Thus this section consists of one or more words that will be repeated or corrected (in case of Repetition or Correction) or abandoned completely (in case of a False Start). It is often followed by a marker called the \textit{interruption point} which is the point at which the speaker realizes that they have made a mistake. The interregnum consists of acknowledgment words that the previous utterance may not be correct. This part consists of an editing term, a non-lexicalized filler pause like "uh" or "um", discourse markers like "well", or "you know", or interjections. Interregnum is followed by the repair, which consists of words spoken to correct previous errors. Words from the reparandum are finally corrected or repeated (in case of Repetition or Correction), or a completely new sentence is started (in case of False Start) in the repair section. In many cases, the interruption point and interregnum may be a simple pause in utterance and thus can be empty in the structure. Figure \ref{fig:dc-example} illustrates the surface structure of disfluent utterances through an example. 

There are six types of disfluencies encountered in real life - Filled Pause, Interjection, Discourse Marker, Repetition or Correction, False Start, and Edit. This section describes each type of disfluency and gives some examples in English.

\subsubsection{Filled Pause}

Filled pauses consist of utterances that have no semantic meaning. 

Example - What about the \textbf{uh} party we have to go to?

\subsubsection{Interjection}

Interjections are similar to filled pauses, but their inclusion in sentences indicates affirmation or negation. 

Example - \textbf{Ugh}, what a night it has been!

\subsubsection{Discourse Marker}

Discourse markers help the speaker begin a conversation or keep a turn while speaking. Just like filled pauses and interjections, these words do not add semantic meaning to the sentence. 

Example - \textbf{Well}, we are going to the party.

\subsubsection{Repetition or Correction}

This disfluency type covers the repetition of certain words in the sentence and correcting words that were incorrectly uttered. 

Example - If I \textbf{can't} don't go to the party today, it is not going to look good. 

\subsubsection{False Start}

False starts occur when a previous chain of thought is abandoned, and a new idea is begun. 

Example - \textbf{Tuesdays don't work for me}, how about Wednesday?

\subsubsection{Edit}

The Edit disfluency type refers to the set of words that are uttered to correct previous statements. 

Example - We need two tickets, \textbf{I'm sorry}, three tickets for the flight to New York.


\subsection{Model Architecture}
\label{sec:app-model}

\subsubsection{wav2vec 2.0 (Referred from Section \ref{met:asr})}
\label{sec:app-asr}

The wav2vec 2.0 model \cite{baevski} consists of a multi-layer convolutional feature encoder that acts on the raw audio input to generate speech representations {${z_1, z_2, \dots, z_T}$} and quantized targets $q_1,q_2,\dots,q_T$ for self-supervision. These representations are fed into a Transformer model \cite{NIPS2017_3f5ee243} to learn context representations $c_1,c_2,\dots,c_T$ which capture the information contained in the entire sequence.  The masked learning objective hides certain time steps in the speech representation space and the objective during training is to predict the quantized targets for these time steps.

\subsubsection{Neural Machine Translation}
\label{sec:app-mt-model}
We used the Transformer \cite{NIPS2017_3f5ee243} architecture for all the NMT models in our experiments. The transformer model has 6 encoder layers and 6 decoder layers. The number of encoder attention heads is 8 and the number of decoder attention heads is 8. The encoder and decoder embedding dimensions are 512. The encoder and decoder feed-forward layer dimensions are 2048. The number of hyperparameters of the Transformer model that we used is 75M.

\subsection{Training Details}
\label{sec:app-train}

\subsubsection{Automatic Speech Recognition (Referred from Section \ref{sec:models})}
\label{sec:app-asr-train}

The ASR architecture we use is based on the wav2vec 2.0 \cite{baevski} framework and consists of 12 blocks of model dimension 768 and 8 attention blocks. Audio samples are re-sampled to 16K KHz and cropped to 250,000 audio frames with a dropout of 0.1. This base model is pre-trained on unlabelled speech data for almost 300K iterations starting with a learning rate of 5e-1. We optimize the pre-training loss function using Adam. During finetuning, a fully connected layer is added after the transformer block for character-level prediction. For our experiments, we use a pretrained transformer encoder so that a limited amount of data is only used to finetune the weights of the outer fully connected layer. 

\subsubsection{Disfluency Correction (Referred from Section \ref{sec:models})}
\label{sec:app-dc-train}

We use the MuRIL \cite{khanuja_muril} transformer model (muril-base-cased) from Hugging Face for our DC experiments. MuRIL consists of a BERT base encoder model pretrained on textual data in 17 Indian languages for the masked language modeling and translated language modeling objectives. Training is performed for 1M steps with a maximum sequence length of 512 and a global batch size of 4096. The AdamW optimizer is used with a learning rate of 5e-4. The final model has 236M parameters. We utilize this pretrained checkpoint and finetune it for disfluency correction by adding a subword token classifier on top of the encoder. For each subword identified by the MuRIL tokenizer, the model predicts if the token is disfluent or fluent.

\subsubsection{Neural Machine Translation (Referred from Section \ref{sec:models})}
\label{sec:app-mt-train}
We used the Indic NLP library for preprocessing the Indic language data and  Moses for preprocessing the English language data. For Indic languages, we normalize and tokenize the data. For English, we lowercase and tokenize the data. We use the byte pair encoding \cite{sennrich-etal-2016-neural} technique to convert the words in the data into subwords. We perform byte pair encoding with 24,000 merge operations on the dataset. We used the fairseq library for training all Transformer based NMT models in all our experiments. We used the Adam optimizer with beta values of 0.9 and 0.98. We used the inverse square root learning rate scheduler with 4000 warm-up updates and a learning rate of 5e-4. The dropout probability value used was 0.1. We used label-smoothed cross-entropy loss with a label-smoothing value of 0.1. The batch size used was 4096 tokens. We train all the models for 200,000 steps and pick the models that give the best loss on the validation set as the final model. We train all our models on a single Nvidia A100 80GB GPU.

\subsubsection{Text-to-Speech}
\label{sec:app-tts-train}

The Forward Tacotron model is sensitive to learning rate scheduling during training. For every language, learning rates were experimentally determined. While training the Tacotron model for extracting alignments, learning rates had to be changed after fixed optimization steps. This process took only 25K steps since we only train for attention alignments. Since we use only around 5 hours of data in each language, we reduce the total optimization steps from 300K to 40K which was sufficient for model convergence.

\subsection{Results (Referred from Section \ref{sec:results})}
\label{sec:app-results}

\subsubsection{Automatic Speech Recognition (Referred from Section \ref{sec:models})}
\label{sec:app-asr-results}

Table \ref{app-tab:res-asr} provides details about the experimentation we perform with different iterations of the wav2vec 2.0 architecture. We also use the Whisper-small model for comparison. Our experiments show that noisy finetuning by training baseline systems with synthetically injected noisy data improves recognition accuracy in Indian English and Hindi. A similar experiment for the Whisper architecture will be conducted in future work.

\subsubsection{Disfluency Correction (Referred from Section \ref{sec:models})}
\label{sec:app-dc-results}

Due to the lack of work in Indian languages disfluency correction, we report the performance of our models compared to a strong baseline with real labeled data. We observe that adding synthetic data in complex disfluencies such as Repetitions and False Starts improves both the precision and recall of our models (Table \ref{app-tab:res-dc}).

\subsubsection{Machine Translation (Referred from Section \ref{sec:models})}
\label{sec:app-mt-results}
Table \ref{app-tab:res-mt} shows the BLEU scores of the best-performing NMT models for all the language pairs. These NMT models are used in the SSMT pipeline. We compute the BLEU scores using the sacrebleu \cite{post-2018-call} library.

\subsubsection{Text To Speech}
\label{sec:app-tts-results}

To compare the performance of our model with a strong baseline, we train the autoregressive Tacotron 2 architecture with the dataset we use to train our Forward Tacotron model. Instead of training from scratch, we use a transliteration module to convert Hindi and Marathi sentences in Devanagari to Roman characters \cite{9342071}. This jumpstarts the training since the model finetunes its grapheme to phoneme embedding without learning them for a new script. TTS systems are evaluated using subjective surveys of speech outputs measuring the scores (out of 5.0) for two parameters - Audio Quality (AQ) and Interpretability (I). The Mean Opinion Score (MOS) is calculated as an average of these two metrics. Table \ref{app-tab:res-tts} shows the results of our TTS system evaluation.

\end{document}

%% file: 1-abstract.tex
\begin{abstract}
In this work, we present our deployment-ready Speech-to-Speech Machine Translation (SSMT) system for English-Hindi, English-Marathi, and Hindi-Marathi language pairs. We develop the SSMT system by cascading Automatic Speech Recognition (ASR), Disfluency Correction (DC), Machine Translation (MT), and Text-to-Speech Synthesis (TTS) models. We discuss the challenges faced during the research and development stage and the scalable deployment of the SSMT system as a publicly accessible web service. On the MT part of the pipeline too, we create a Text-to-Text Machine Translation (TTMT) service in all six translation directions involving English, Hindi, and Marathi. To mitigate data scarcity, we develop a LaBSE-based corpus filtering tool to select high-quality parallel sentences from a noisy pseudo-parallel corpus for training the TTMT system. All the data used for training the SSMT and TTMT systems and the best models are being made publicly available. Users of our system are (a) Govt. of India in the context of its new education policy\footnote{\label{foot-nep}\url{https://www.education.gov.in/sites/upload_files/mhrd/files/NEP_Final_English_0.pdf}} (NEP), (b) tourists who criss-cross the multilingual landscape of India, (c) Indian Judiciary where a leading cause of the pendency of cases (to the order of 10 million as on date) is the translation of case papers, (d) farmers who need weather and price information and so on. We also share the feedback received from various stakeholders when our SSMT and TTMT systems were demonstrated in large public events.
\end{abstract}

%% file: 2-intro.tex
\section{Introduction} \label{sec:intro}
Speech-to-Speech Machine Translation (SSMT) is the task of automatically translating spoken utterances of one language into spoken utterances of another language; similarly, Text-to-Text Machine Translation (TTMT) is the task of translating the text of one language into that of another. SSMT and TTMT have many important applications.

India's new National Education Policy\textsuperscript{\ref{foot-nep}} (NEP) includes a new language policy that states the preferred medium of instruction should be the mother tongue, local or regional language till Class 5 or even Class 8. Our system will aid the teachers by providing a platform for the live translation of lectures in vernacular Indian languages. The tourism industry in India handled about 700 million tourists (677M domestic and 7M foreign) in 2021\footnote{\url{https://tourism.gov.in/market-research-and-statistics}}, and it requires translation on a day-to-day basis. About 43.5 million\footnote{\url{https://njdg.ecourts.gov.in/njdgnew/?p=main}} cases are pending in Indian Judiciary. A major reason for this is that documents, such as FIR, police charge sheets, and also court judgments are in vernacular languages, and for these appeals to be listed for hearing in higher courts, these documents need to be translated into English\footnote{\url{https://theprint.in/judiciary/sc-will-now-translate-daily-orders-and-judgments-into-9-languages-using-ai-tools/429597/}}. TTMT is crucial in Indian Judiciary. The healthcare sector is also a major application of translation as translating healthcare information of diseases such as Covid-19\footnote{\url{https://www.wired.com/story/covid-language-translation-problem/}} plays an important role in controlling the spread of diseases. The SSMT system can also be used by doctors to efficiently communicate the diagnosis to patients in their native language.

Our English-Hindi, English-Marathi, and Hindi-Marathi SSMT system is developed by cascading the Automatic Speech Recognition (ASR), Disfluency Correction (DC), Machine Translation (MT), and Text-to-Speech (TTS) models. The system is efficiently deployed on our servers to serve multiple concurrent users with very low latency. We have also created a public web service through which users can easily access our SSMT and TTMT systems. We have collected feedback on our SSMT system from various stakeholders.


Our contributions are:
\begin{compactenum}
    \item Deployment of scalable speech-to-speech and text-to-text machine translation systems for English-Hindi, English-Marathi, and Hindi-Marathi (salient linguistic properties of the languages are mentioned in \textbf{Appendix \ref{app:linguistic}}) language pairs.
    \item Demonstrating a corpus filtering toolkit that can be used to extract high-quality parallel corpus from the noisy pseudo parallel corpus and misaligned parallel corpus.
\end{compactenum}


%% file: 3-related-work.tex
\section{Related Work} \label{sec:rel_work}
We find two popular SSMT approaches in the literature. Recently an end-to-end speech translation system~\cite{lee-etal-2022-direct} that uses a single neural network is developed. But, it requires a huge amount of high-quality speech-to-speech parallel corpus. We follow the pipeline-based approach, which does not require the speech-to-speech parallel corpus, and involves connecting different components in a cascade to form the SSMT pipeline~\cite{bahar-etal-2020-start}. The literature for each component of our SSMT system is discussed below.


Hidden Markov Models improved the performance of traditional ASR \cite{5740583} by associating every phoneme of a given language to an HMM model with transition and emission probabilities obtained from the training corpus \cite{SIG-004}. Over the last 25 years, the amount of labeled training data has increased in many languages, which has allowed deep learning-based systems to leverage recorded and transcribed speech. From replacing acoustic models in traditional ASR \cite{5740583} to functioning as an end-to-end speech recognition system \cite{deepspeech,7472621,graves_asr,rnnt_asr}, deep learning has played a pivotal role in transforming ASR across many languages. Appendix \ref{sec:asr-background} describes the mathematical formulation and evaluation metrics of ASR. 


Disfluency correction is an essential pre-processing step to clean disfluent sentences before passing the text through downstream tasks like machine translation \cite{rao-etal-2007-improving,5494999}. There are three main approaches in developing DC systems: noisy channel-based techniques \cite{honal_dc,jamshid-lou-johnson-2017-disfluency}, parsing-based techniques \cite{honnibal-johnson-2014-joint, jamshid-lou-johnson-2020-improving}, and sequence tagging-based techniques \cite{hough_dc,ostendorf_dc}. Synthetic disfluent data generation by infusing disfluent elements in fluent sentences has received attention recently to compensate for the lack of annotated data in low resource languages \cite{passali-etal-2022-lard,saini-etal-2020-generating}. Appendix \ref{sec:dc-background} discusses the surface structure of disfluencies and their various types. 

NMT models are \textit{data hungry} \cite{cho-etal-2014-learning, sutskever2014sequence, 38ed090f8de94fb3b0b46b86f9133623, NIPS2017_3f5ee243}. \citet{kim-etal-2019-pivot} proposed pivot-language-based transfer learning techniques for NMT in which the encoder and decoder of source-pivot and pivot-target NMT models are used to initialize the source-target model. \citet{sennrich-etal-2016-improving} proposed the backtranslation technique in which the synthetic data is created by translating monolingual data. \citet{sen2021neural} proposed the phrase pair injection technique in which source-target phrase pairs generated from the source-target parallel corpus using SMT are augmented with source-target parallel corpus. The bad-quality phrase pairs can be filtered out using LaBSE-based \cite{feng-etal-2022-language} corpus filtering techniques \cite{batheja-bhattacharyya-2022-improving}.


Text-to-Speech generates intelligible and natural-sounding speech using only the input text prompt. Statistical frameworks model speech synthesis through a transition network optimizing a defined cost function by concatenating different phonemes \cite{541110}. Deep learning-based systems directly learn the mapping between phoneme-sequence and mel spectrograms, which accurately represent acoustic and prosodic information. WaveNet \cite{vandenoord16_ssw} and Tacotron \cite{8461368} are examples of such black box architectures.

%% file: 4-method.tex
\section{Method} \label{sec:method}

\begin{figure*}[ht!]
\centering
\includegraphics[width=160mm]{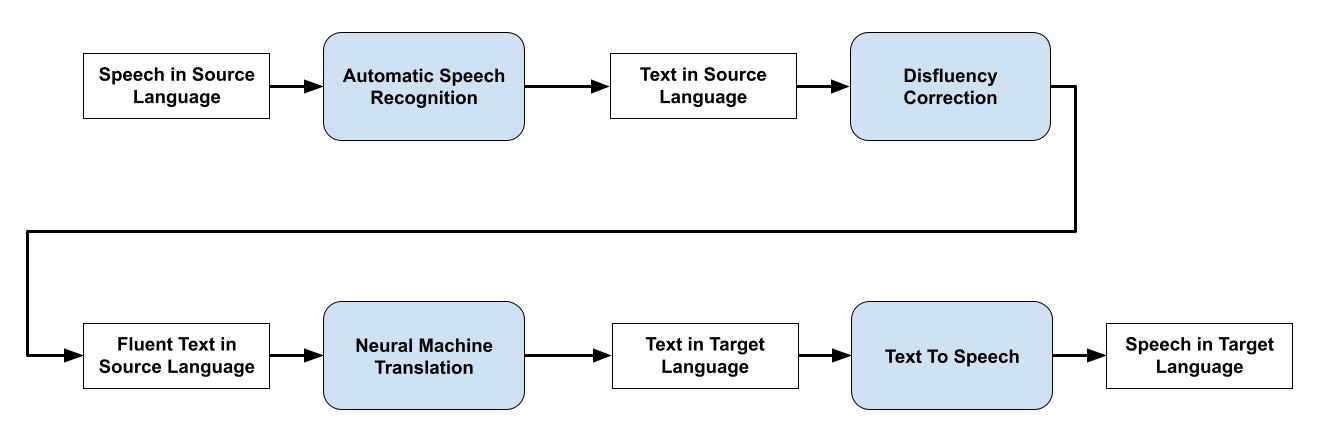}
\caption{Speech-to-Speech Machine Translation pipeline} \label{fig:ssmt}
\end{figure*}

In this section, we discuss in detail all components and their linking that form the SSMT pipeline.

\subsection{System Overview}
The SSMT system consists of a cascade of four components as demonstrated by Figure \ref{fig:ssmt}: The input speech is passed to the ASR system, which transcribes the input speech in the source language. This transcription may also contain disfluencies from the speech, which are removed using the DC system. This source language text is translated into the target language text using the MT system. The TTS system then generates the corresponding target language speech.

\vspace{-5pt}

\subsection{Automatic Speech Recognition}
\label{met:asr}

Recent deep learning techniques utilize unlabelled speech data using self-supervision and masked language modeling \cite{Schneider2019wav2vecUP,10.5555/3495724.3496768}. These methods pre-train large transformer models on vast quantities of unlabelled speech data to learn high-quality speech representations, followed by finetuning on limited labeled data to generate transcripts of a spoken utterance. \citet{ruder-etal-2019-unsupervised} extends this technique for multilingual training facilitating finetuning in many low-resource languages. More recently, the Whisper ASR system \cite{Radford2022RobustSR} promises robust speech recognition leveraging vast quantities of weakly supervised parallel data to train large transformers for speech recognition and translation. Our ASR system is inspired by \citet{DBLP:journals/corr/abs-2107-07402}, which pre-trains a wav2vec 2.0 model (Appendix \ref{sec:app-asr}) using unlabelled speech data in Indian languages followed by finetuning on labeled data in English and Hindi respectively. We further train this model on Signal to Noise Ratio (SNR) modulated audio samples to make the transcription quality more robust in noisy environments. 

\vspace{-5pt}

\subsection{Disfluency Correction}


Our disfluency correction system is based on \citet{kundu-etal-2022-zero} which trains a large multilingual transformer model using real and synthetic data in English and Hindi. We use Google's MuRIL transformer \cite{khanuja_muril} model since it has better representations for Indian languages and is trained on annotated data for token classification.

\subsection{Machine Translation}
The TTMT is a task of automatically translating source language text into target-language text. In this section, we explain the data-preprocessing and the system development phases.
\paragraph{LABSE-based Corpus Filtering}
The task of Parallel Corpus Filtering aims to provide a scoring mechanism that helps extract good-quality parallel corpus from a noisy pseudo-parallel corpus. \citet{labse} proposed the LaBSE model, which is a multilingual sentence embedding model trained on 109 languages, including some Indic languages. \citet{batheja-bhattacharyya-2022-improving} proposed a combined approach of Phrase Pair Injection and LaBSE-based corpus filtering that helps improve the quality of MT systems. We have developed a \textbf{LaBSE-toolkit} that performs the following three tasks:
\begin{compactenum}
	\item \textbf{Assign quality scores to the Parallel Corpus:} We use the LaBSE model to generate sentence embeddings for the source and target sentences. Then, we compute the cosine similarity between the source and target sentence embeddings.

\item \textbf{Extract high-quality Parallel Corpus from the noisy Pseudo-Parallel Corpus:} We extract high-quality Parallel Corpus from the noisy Pseudo-Parallel Corpus based on a threshold quality score value.

\item \textbf{Sentence Alignment in misaligned Parallel Corpus:} We compute the quality score for each source and target sentence and match the sentence pair with the maximum quality score value.
\end{compactenum}

\paragraph{Neural Machine Translation (NMT)} 
Neural Machine Translation systems which use deep learning models have shown great performance in the task of translation. But NMT models are \textit{data hungry} that is they require large amounts of training corpus for training. This limitation creates a hurdle in training good-quality NMT models for low-resource language pairs. In such cases, a related pivot language can be used as an assisting language for the low-resource language pair. 
The naive cascade pivoting \cite{de2006catalan} approach suffers from the problems of double decoding time and propagating errors. To avoid these problems we need to train a single source-target NMT model that utilizes the resources of the pivot language. We use the pivot-based transfer learning approach in which we first train a source-pivot and pivot-target NMT model. Then we initialize the source-target model NMT with the encoder and decoder of the source-pivot and pivot-target NMT models. Then we finetune the source-target NMT model on the source-target parallel data.
\par
We use the transformer architecture for all the NMT models. The models are trained using the fairseq~\cite{ott-etal-2019-fairseq} library. We use the ctranslate2\footnote{\url{https://github.com/OpenNMT/CTranslate2}} library for fast and efficient inference of the NMT models in deployment.

\subsection{Text-To-Speech}



Models like WaveNet \cite{vandenoord16_ssw} and Tacotron \cite{8461368} are autoregressive speech synthesis frameworks that predict speech frames one-time step at a time. Since such recurrent prediction frameworks result in slow inference during deployment, non-autoregressive deep learning models like FastSpeech \cite{10.5555/3454287.3454572} and Forward Tacotron\footnote{\url{https://github.com/as-ideas/ForwardTacotron}} were developed which generate speech frames in a single run. Our TTS model is an adaptation of the Forward Tacotron architecture trained on high-quality speech synthesis datasets.

%% file: 5-expt-setup.tex
\section{Experimental Setup} \label{sec:expt_setup}
In this section, we discuss the datasets and models that we used for all the experiments. 
\subsection{Dataset}
\begin{table}[htp]
\centering
\begin{tabular}{c c}
\hline 
\textbf{Language Pair} &\textbf{\# of Sentence Pairs}\\
\hline
\hline 
English-Hindi & 9.4M \\
English-Marathi & 6.2M \\ 
Hindi-Marathi & 2.55M \\ \hline
\end{tabular}
\caption{Dataset Statistics for the task of NMT}
\label{lab:corpus-mt}
\end{table}
\begin{table}[htp]
\centering
\begin{tabular}{c c c}
\hline 
\textbf{Language} &\textbf{ASR} & \textbf{TTS}\\
&(\# of hours) & (\# of hours)\\
\hline
\hline 
English&  25.57 & NA \\
Hindi& 24.92 & 4.57 \\ 
Marathi& NA  & 4.82\\ \hline
\end{tabular}
\caption{Dataset Statistics for the task of ASR and TTS}
\label{lab:corpus-asr}
\end{table}
\begin{table*}[h]
\centering
\begin{tabular}{p{6cm}p{9cm}}
\hline
\textbf{Stakeholder} & \textbf{Feedback} \\
\hline
\hline
Co-founder of a voice services start-up & We want to build a conversation bot for payment through UPI123Pay. We would like to integrate MT with our system. \\
\hline
Doctors from a Mental Health Hospital & SSMT can be used in tele-counseling for national telementor health programs. \\
\hline
An employee of an agricultural technology company & We would like to explore the integration of the SSMT with our application for farmers. \\
\hline
An employee of a stock trading startup & Interested in collaboration. \\
\hline
An employee of an electric vehicle charging station solutions company & Use SSMT in providing voice support for EV charging stations. \\
\hline
An employee of an e-commerce company & Very nice initiative and approach. Requirement of Translation of e-commerce website. \\
\hline
\end{tabular}
\caption{Feedback received from potential stakeholders of our SSMT and TTMT systems.}
\label{tab:feedback}
\end{table*}

\noindent \textbf{ASR} \quad  We choose the CommonVoice dataset where speakers read text prompts and record audio using Computers, mobiles, etc., without professional recording instruments \cite{ardila-etal-2020-common}. This method infuses noise due to the speaker's environmental conditions. We further augment noise to a part of this dataset to retrain the baseline model \cite{reddy-noise}. Due to resource constraints, we use approximately 25 hours of aligned speech data for English and Hindi, with 20 hours for training and 5 hours for testing. For evaluating English ASR, we use data from the test set of NPTEL2020 - Indian English speech dataset\footnote{\url{https://github.com/AI4Bharat/NPTEL2020-Indian-English-Speech-Dataset}}. 
\\
\textbf{DC} \quad Our DC models are trained on Switchboard, the largest English annotated disfluency correction dataset \cite{Godfrey1992SWITCHBOARDTS}. For English, we combine the Switchboard dataset with labeled synthetic disfluent data from \citet{passali-etal-2021-towards} to create an equitable distribution of samples from all disfluency types for training and testing. For Hindi, we utilize parallel disfluent-fluent sentences from Switchboard and augment them with synthetic disfluent sentences in Hindi created from transcribed fluent utterances by rule-based disfluency injection methods \cite{kundu-etal-2022-zero}. The Hindi DC model is evaluated on a gold standard dataset created by human-transcribed speech samples from YouTube podcasts and interviews. 
\\
\textbf{MT} \quad We have used various open-source parallel corpora such as Samanantar \cite{ramesh-etal-2022-samanantar}, Anuvaad, LoResMT \cite{mtsummit-2021-technologies} workshop dataset, ILCI \cite{jha-2010-tdil}, and Spoken Tutorial dataset. We also create a parallel corpus for En-Mr of 120K sentences with the help of translation startups. The detailed dataset statistics of the parallel corpora used are mentioned in Table \ref{lab:corpus-mt}.
\\
\textbf{TTS} \quad To train our TTS models, we use the IndicTTS dataset \cite{baby_indictts} containing noise-free read speech. The corpus consists of 4.57 and 4.82 hours of speech data in Hindi and Marathi, respectively. We re-sample speech files to 22.05 kHz and remove terminal silence. We use the e-speak phonemizer \cite{Bernard2021} to convert graphemes to phonemes during training and inference. Dataset details are specified in Table \ref{lab:corpus-asr} and training details are mentioned in Appendix \ref{sec:app-tts-train}.

\subsection{Models}
\label{sec:models}
\textbf{ASR} \quad We use the Hugging Face checkpoint of Vakyansh English and Hindi ASR models and finetune it on the noisy dataset we created. Our experiments use a 12 GB NVIDIA GeForce RTX GPU, which significantly reduces training time. For more details, please refer to Appendix \ref{sec:app-asr-train} and \ref{sec:app-asr-results}. 
\\
\textbf{DC} \quad The MuRIL \cite{khanuja_muril} checkpoint from Hugging Face is used to finetune our DC models. We use the transformers package to train this BERT-based encoder for binary token classification, i.e., the model needs to predict whether each word in the sentence is disfluent or fluent. Training details and baseline comparisons have been provided in Appendix \ref{sec:app-dc-train} and \ref{sec:app-dc-results} respectively. 
\\
\textbf{MT} \quad The NMT models are based on the Transformer architecture. We use the fairseq library to train all models. The detailed model architecture and training details are mentioned in Appendix \ref{sec:app-mt-model} and \ref{sec:app-mt-train}. The BLEU scores of the best-performing NMT models for all language pairs on different test sets containing sentences from different domains are mentioned in Appendix \ref{sec:app-mt-results}.
\\
\textbf{TTS} \quad The forward tacotron architecture replaces 12 memory-consuming self-attention transformer layers of FastSpeech \cite{ren_fastspeech} with the recurrent prediction framework from \cite{shen-tacotron2}. The autoregressive nature of training is removed by adding a length regulator to predict mel spectrograms in a single pass. The length regulator uses a separately trained duration predictor model that expands phoneme embeddings based on predicted duration. Finally, a CARGAN \cite{morrison2022chunked} based vocoder converts the mel spectrograms into audio files.

%% file: 6-results.tex
\section{Results} \label{sec:results}
\begin{table}[h]
\centering
\begin{tabular}{lrrr}
\hline 
\textbf{Metrics} &\textbf{En-Hi} & \textbf{En-Mr} & \textbf{Hi-Mr}\\
\hline
\hline
TQ & 4.43 & 4.11 & 4.08 \\ 
SQ & 4.64 & 4.53 & 4.63\\ 
I & 4.60 & 4.51 & 4.87\\ 
\hline
\end{tabular}
\caption{Human Evaluation Scores of the SSMT system. The number of participants in the survey was 101. En: English, Mr: Marathi, Hi: Hindi, \textbf{TQ}: Translation Quality, \textbf{SQ}: Speech Quality, \textbf{I}: Interpretability; All scores are measured out of 5.0}
\label{lab:ssmt-survey-results}
\end{table}

In this section, we discuss the performance of our scalable SSMT system. Since there are no automatic evaluation metrics to evaluate an SSMT system, we perform subjective evaluation by conducting a widescale survey. We asked 101 participants to rate five samples per language pair on three key performance indicators (KPIs): Translation Quality (TQ), Speech Quality (SQ), and Interpretability (I) on a scale of 0 to 5. The participants were asked to listen to human-generated source language speech and SSMT-generated target language speech. Results of this survey are described in Table \ref{lab:ssmt-survey-results}. 

Our SSMT system performs well on all three KPIs for all three translation directions.
The English-Hindi SSMT system demonstrates the highest translation quality by receiving a TQ score of 4.43. For all three directions, the Speech Quality and Interpretability scores are more than 4.5 out of 5, which shows that the TTS system produces good-quality speech output. The performance of our individual ASR, DC, MT, and TTS systems are described in Appendix \ref{sec:app-results}. The BLEU scores of the MT models represent the quality of our TTMT system.


%% file: 7-feedback.tex
\section{Stakeholder Feedback} \label{sec:feedback}
We showcased our SSMT and TTMT systems at large public events to get exposure and gather feedback. Various potential stakeholders like industry personnel, government officials, professors, students, and individuals interacted with our systems. Table \ref{tab:feedback} lists remarks made by a few of them.

Along with an appreciation for our solutions for their robustness, we also received valuable suggestions for improving our systems and extending our work. It led to improvements like reduced latency for both the MT systems, the addition of features including an automatic pause detection in the SSMT and an automatic language detection in the TTMT, enhanced UI, \textit{etc.} Discussions with a diverse set of people educated us about unique, sought-after applications of the SSMT and TTMT systems. The eagerness expressed by people from various backgrounds to adopt our systems for their specific use cases has shown new directions to extend our work.

%% file: 8-deploy.tex
\section{Deployment} \label{sec:deploy}
\begin{table}[h]
\centering
\begin{tabular}{p{2cm}p{2cm}p{2cm}}
\hline
\textbf{Number of Concurrent Users} & \textbf{Deployed System} & \textbf{Baseline System} \\
 & (in msec.) & (in msec.) \\
\hline
\hline

50 & 1,600 & 12,727 \\
100 & 2,200 & 25,715 \\
500 & 3,300 & 27,214 \\
1000 & 4,400 & 60,043 \\
\hline
\end{tabular}
\caption{Median response times (in milliseconds) of the SSMT system for different numbers of concurrent users. We compare the median repose times of the deployed system with the baseline system. The \textbf{baseline system} consists of a single SSMT pipeline running on Nvidia RTX 2080Ti GPU. The \textbf{deployed system} consists of 104 SSMT pipelines running on the Nvidia DGX A100 machine which consists of 8 Nvidia A100 80GB GPUs.}
\label{tab:load-testing}
\end{table}
We have deployed the SSMT system as a web service using ReactJS for the frontend and FastAPI for the backend. The frontend has the functionality to record the input speech and make API calls to the backend with the input speech as a payload. The backend runs the SSMT pipeline (ASR, DC, MT, and TTS models) and the input speech is passed through the SSMT pipeline to generate the output speech. The outputs of the ASR (input transcript), MT (output sentence), and TTS (output speech) are sent to the frontend which displays the output. 
\par
We have deployed the ASR, DC, MT, and TTS models on the Nvidia DGX A100 machine. The Nvidia DGX A100 machine has 8 Nvidia A100 GPUs and each GPU has 80GB of GPU memory. Each SSMT pipeline (which consists of ASR, DC, MT, and TTS models) occupies a space of around 6GB on GPU memory. In order to efficiently support multiple users at a time and utilize all the GPU memory, we deploy multiple SSMT pipelines on the machine. On each GPU we deploy 13 SSMT pipelines and in total, we deploy 104 SSMT pipelines across the 8 GPUs in the Nvidia DGX A100 machine. We also perform load testing of the SST system using the Locust tool. Table \ref{tab:load-testing} shows the median response times of the SSMT system for different numbers of concurrent users. For 1000 concurrent users the SSMT system has a median response time of 4.4 sec.

%% file: 9-conclusion.tex
\section{Conclusion and Future Work} \label{sec:conclusion}

In this work, we develop Speech-to-Speech Machine Translation (SSMT) and Text-to-Text Machine Translation (TTMT) systems for English, Hindi, and Marathi languages. For SSMT, we follow the cascade-based approach that includes Automatic Speech Recognition (ASR), Disfluency Correction (DC), Machine Translation (MT), and Text-to-Speech Synthesis (TTS) components. We also develop the LaBSE-based parallel corpus filtering tool to extract high-quality parallel sentences from a noisy pseudo-parallel corpus for training the TTMT system. We deploy our SSMT and TTMT systems to be scalable so that multiple concurrent users are able to access them with very low latency. We test our systems in the real world and gathered invaluable feedback from various stakeholders that helped further improve our systems.
\par
We are actively working towards incorporating more Indian languages into our SSMT and TTMT systems. We are also focusing on a more robust ASR system that is resistant to different accents, dialects, and noises in speech. Future work in TTS will incorporate more Indian languages as well as modifications in our architecture to generate more human-sounding speech. We are working on deploying our SSMT and TTMT systems on multiple GPU clusters to increase the number of concurrent users that can be served maintaining low latency.

%% file: 10-limitations.tex
The SSMT system brings certain limitations due to design decisions and the inherent complexity of the task at hand. Some of the major ones are:

\begin{compactenum}
    \item Our SSMT system is a cascade of multiple components. This architecture implies that errors from any single component are propagated through the pipeline. Hence each component in the system must be robust to ensure high-quality output.
    \item The cascade-based SSMT approach has multiple models which increase the computational requirements and latency of the system. 
    \item End-to-End SSMT systems can potentially capture all the information in the input speech signal, like emotions and accents of speakers. These abilities are missing in the cascade-based SSMT approach that we follow.
    
\end{compactenum}